\documentclass{article}

\usepackage{arxiv}

\usepackage[utf8]{inputenc} 
\usepackage[T1]{fontenc}    
\usepackage{hyperref}       
\usepackage{url}            
\usepackage{booktabs}       
\usepackage{amsfonts}       
\usepackage{nicefrac}       
\usepackage{microtype}      
\usepackage{lipsum}
\usepackage{graphicx}
\graphicspath{ {./images/} }
\usepackage{amsmath}
\usepackage{algorithmic}
\usepackage{algorithm}
\usepackage{booktabs} 
\usepackage{multirow}
\usepackage{tikz}   
\usetikzlibrary{shapes, arrows, positioning, fit, calc}

\title{FairViT-GAN: A Hybrid Vision Transformer with Adversarial Debiasing for Fair and Explainable Facial Beauty Prediction}

\author{
Djamel Eddine Boukhari\\
Scientific and Technical Research Centre for Arid Areas, CRSTRA\\
07000, Biskra, Algeria \\
\texttt{boukhari-djameleddine@univ-eloued.dz} \\
}

\begin{document}
\maketitle
\begin{abstract}
	Facial Beauty Prediction (FBP) has made significant strides with the application of deep learning, yet state-of-the-art models often exhibit critical limitations, including architectural constraints, inherent demographic biases, and a lack of transparency. Existing methods, primarily based on Convolutional Neural Networks (CNNs), excel at capturing local texture but struggle with global facial harmony, while Vision Transformers (ViTs) effectively model long-range dependencies but can miss fine-grained details. Furthermore, models trained on benchmark datasets can inadvertently learn and perpetuate societal biases related to protected attributes like ethnicity. To address these interconnected challenges, we propose \textbf{FairViT-GAN}, a novel hybrid framework that synergistically integrates a CNN branch for local feature extraction and a ViT branch for global context modeling. More significantly, we introduce an adversarial debiasing mechanism where the feature extractor is explicitly trained to produce representations that are invariant to protected attributes, thereby actively mitigating algorithmic bias. Our framework's transparency is enhanced by visualizing the distinct focus of each architectural branch. Extensive experiments on the SCUT-FBP5500 benchmark demonstrate that FairViT-GAN not only sets a new state-of-the-art in predictive accuracy, achieving a Pearson Correlation of \textbf{0.9230} and reducing RMSE to \textbf{0.2650}, but also excels in fairness. Our analysis reveals a remarkable \textbf{82.9\% reduction in the performance gap} between ethnic subgroups, with the adversary's classification accuracy dropping to near-random chance (52.1\%). We believe FairViT-GAN provides a robust, transparent, and significantly fairer blueprint for developing responsible AI systems for subjective visual assessment.
\end{abstract}


\section{Introduction}
Facial Beauty Prediction (FBP) is a long-standing challenge in computer vision that seeks to computationally model the human perception of facial attractiveness \cite{b1, b2}. It has diverse applications ranging from photo retouching and virtual try-ons to medical aesthetics and psychological research\cite{b3}. Early approaches relied on geometric ratios and handcrafted features, but the advent of deep learning, particularly Convolutional Neural Networks (CNNs)\cite{b4}, has dramatically improved prediction performance by automatically learning hierarchical features from data \cite{b5}.

Despite their success, current FBP models face three critical challenges. First, \textbf{architectural limitations}: CNN-based models are inherently strong at capturing local patterns like texture and edges but their limited receptive fields struggle to model the holistic, long-range spatial relationships between facial components (e.g., the harmony between eyes, nose, and mouth) \cite{b6}. Recently, Vision Transformers (ViTs) \cite{b7}have emerged as a powerful alternative, demonstrating superior ability in modeling global context by treating an image as a sequence of patches \cite{b8}. However, ViTs may not capture the high-frequency details as effectively as CNNs. This suggests that a hybrid approach could yield a more comprehensive feature representation \cite{b9}.

Second, \textbf{algorithmic bias}: FBP models are trained on datasets labeled by human annotators, and they can inherit and even amplify societal biases present in the data \cite{b10}. A model trained on a dataset with a specific demographic majority might perform poorly on minority groups or, worse, develop a biased standard of beauty that favors majority features. This raises significant ethical concerns, limiting the responsible deployment of such technologies \cite{b11, b12}. Addressing this bias is not merely an ethical imperative but also a technical challenge to improve model generalization.

Third, \textbf{lack of transparency}: Most deep learning models for FBP operate as "black boxes," making it difficult to understand the basis for their predictions. This opacity undermines trust and prevents auditing for fairness or diagnosing failures. An explainable model would provide visualizations indicating which facial regions most influence its score, offering valuable insights for both researchers and end-users.

To address these interconnected challenges, we propose a novel framework named \textbf{FairViT-GAN}. Our contributions are threefold:
\begin{itemize}
    \item We design a novel \textbf{hybrid feature extractor} that leverages a CNN branch to learn local texture and detail features, and a ViT branch to learn global shape and structural relationships. These parallel streams are fused to create a rich and comprehensive facial representation.
    \item We introduce an \textbf{adversarial debiasing mechanism}. Our model is trained within a GAN-like framework where a predictor network is pitted against an adversary network. The adversary's goal is to identify a protected attribute (e.g., ethnicity) from the learned feature embedding, while the predictor's goal is to learn an embedding that is not only useful for beauty prediction but also "fools" the adversary. This encourages the model to learn features of beauty that are uncorrelated with the protected attribute.
    \item We incorporate an \textbf{explainability component} by visualizing the attention maps from the ViT branch and applying Grad-CAM to the CNN branch. This provides a dual perspective on the model's focus, highlighting both global and local regions of interest.
\end{itemize}

We validate FairViT-GAN on the large-scale SCUT-FBP5500 benchmark dataset \cite{b10}. Our results show that our model not only sets a new state-of-the-art in prediction accuracy but also demonstrates significantly fairer performance across different demographic subgroups.

The remainder of this paper is organized as follows: Section II reviews related work. Section III details our proposed methodology. Section IV describes our experimental setup, and Section V presents and discusses the results. Finally, Section VI concludes the paper.

\section{Related Work}
\subsection{Deep Learning for Facial Beauty Prediction}
Modern FBP research has been dominated by deep learning. Early deep learning models used standard CNN architectures like AlexNet \cite{b13}, treating the task as a regression problem to predict a continuous beauty score \cite{b14}. Subsequent works introduced more advanced architectures like ResNet \cite{b15} to achieve deeper models and better performance. Multi-task learning has also been explored, where FBP is jointly trained with related tasks like facial landmark detection, gender classification, or age estimation, leveraging shared representations to improve the main task \cite{b16}. However, these models primarily rely on standard CNN backbones and do not explicitly address the bias and explainability issues.

\subsection{Vision Transformers in Facial Analysis}
The Vision Transformer (ViT) \cite{b7} has recently shown great promise in computer vision tasks. By dividing an image into patches and using a Transformer encoder, ViT excels at capturing global context\cite{b17}. In facial analysis, ViTs have been successfully applied to tasks like face recognition and facial expression recognition \cite{b18}. Some studies have applied ViTs to FBP, noting that they tend to focus more on global features like facial contours and skin texture compared to CNNs which focus on local parts like eyes and mouth \cite{b19}. Our work builds on this by proposing a hybrid model, aiming to get the best of both worlds.

\subsection{Bias and Fairness in AI}
The problem of bias in AI systems is a critical area of research \cite{b20}. In facial analysis, it has been shown that commercial face recognition systems can have significantly higher error rates for women and people of color. The cause is often biased training data that underrepresents these groups \cite{b21}. Several techniques have been proposed to mitigate bias, including re-weighting training samples, data augmentation, and adversarial training. Adversarial debiasing, which we adopt, has proven effective in learning representations that are invariant to protected attributes. While fairness has been studied extensively in face recognition, its application to the subjective domain of FBP is still nascent.

\section{Proposed Methodology: FairViT-GAN}
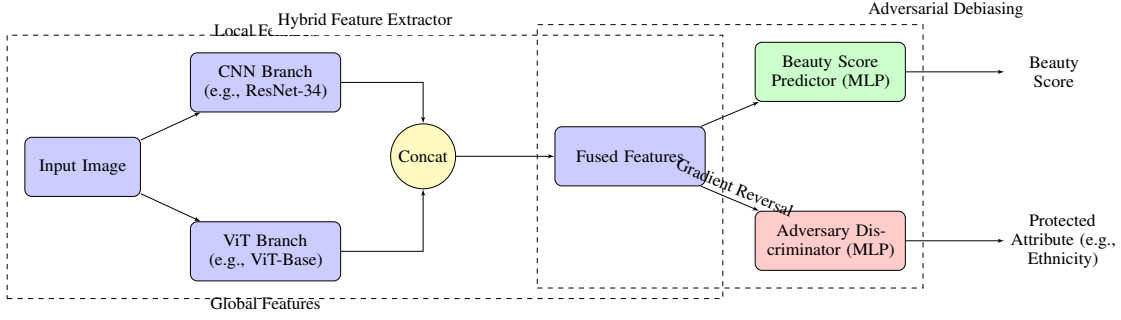
\begin{figure*}[!t]
    \centering
    \resizebox{0.9\textwidth}{!}{%
    \begin{tikzpicture}[
        node distance=1.5cm and 2cm,
        block/.style={rectangle, draw, fill=blue!20, text width=8em, text centered, rounded corners, minimum height=12mm},
        sum/.style={circle, draw, fill=yellow!30, node distance=1cm},
        line/.style={draw, -latex'}
    ]
        \node[block, text width=6em] (input) {Input Image};

        \node[block, above right=0.5cm and 1cm of input] (cnn) {CNN Branch (e.g., ResNet-34)};
        \node[block, below right=0.5cm and 1cm of input] (vit) {ViT Branch (e.g., ViT-Base)};

        \node[above=0.2cm of cnn] (local_label) {Local Features};
        \node[below=0.2cm of vit] (global_label) {Global Features};
        
        \node[sum, right=of cnn, yshift=-1.5cm] (concat) {Concat};

        \node[block, right=of concat] (fused) {Fused Features};

        \node[block, fill=green!20, above right=0.5cm and 1cm of fused] (predictor) {Beauty Score Predictor (MLP)};
        \node[text width=5em, text centered, right=of predictor] (output) {Beauty Score};
        
        \node[block, fill=red!20, below right=0.5cm and 1cm of fused] (adversary) {Adversary Discriminator (MLP)};
        \node[text width=6em, text centered, right=of adversary] (protected) {Protected Attribute (e.g., Ethnicity)};

        \path[line] (input) -- (cnn);
        \path[line] (input) -- (vit);
        \path[line] (cnn) -| (concat);
        \path[line] (vit) -| (concat);
        \path[line] (concat) -- (fused);

        \path[line] (fused) -- (predictor);
        \path[line] (fused) -- node[midway, sloped, above] {Gradient Reversal} (adversary);
        
        \path[line] (predictor) -- (output);
        \path[line] (adversary) -- (protected);
        
        \node[draw, dashed, inner sep=10pt, label={[fill=white]above:Hybrid Feature Extractor}, fit=(input) (cnn) (vit) (concat) (fused)] {};
        \node[draw, dashed, inner sep=10pt, label={[fill=white]above right:Adversarial Debiasing}, fit=(fused) (predictor) (adversary)] {};

    \end{tikzpicture}
    }
    \caption{The overall architecture of our proposed FairViT-GAN model. An input image is processed in parallel by a CNN branch to capture local features and a ViT branch to capture global features. The resulting feature vectors are concatenated and fed into two heads: a predictor for the beauty score and an adversary for the protected attribute. A gradient reversal layer ensures that while the adversary learns to predict the attribute, the feature extractor learns to be invariant to it.}
    \label{fig:architecture}
\end{figure*}

Our proposed FairViT-GAN framework consists of three main components: (1) a hybrid feature extractor for comprehensive facial representation, (2) a score prediction head for the primary regression task, and (3) an adversarial discriminator for debiasing. The overall architecture is depicted in Fig. \ref{fig:architecture}.

\subsection{Hybrid Feature Extractor}
The feature extractor is the backbone of our model. It processes an input image $I$ to produce a feature embedding $f$.
\paragraph{CNN Branch}
We use a truncated ResNet-34, pre-trained on ImageNet, as our CNN branch. The convolutional layers are adept at extracting a rich hierarchy of local features, from simple edges and textures in early layers to more complex object parts in deeper layers. We remove the final fully connected layer and extract the feature vector $f_{cnn} \in \mathbb{R}^{d_{cnn}}$ from the global average pooling layer.
\paragraph{ViT Branch}
We use a standard Vision Transformer, specifically ViT-Base. The input image is first split into a sequence of fixed-size patches. These patches are linearly embedded and then augmented with position embeddings. The resulting sequence of vectors is fed into a Transformer encoder, which consists of multiple self-attention and MLP blocks. The self-attention mechanism allows the model to weigh the importance of all other patches when representing a given patch, thus capturing global context. We use the output embedding corresponding to the special `[CLS]` token as the global feature vector $f_{vit} \in \mathbb{R}^{d_{vit}}$.

Finally, the local and global features are concatenated to form the final fused feature embedding:
\begin{equation}
    f = [f_{cnn} \,;\, f_{vit}]
\end{equation}
where $;$ A denotes concatenation.

\subsection{Beauty Prediction and Adversarial Debiasing}
The fused feature vector $f$ is passed to two separate MLP heads.

\paragraph{Predictor Network} The predictor, $P$, takes $f$ as input and regresses to a continuous beauty score $\hat{y}$. It is a simple MLP with two hidden layers. We train the predictor and the feature extractor using a standard Mean Squared Error (MSE) loss:
\begin{equation}
    \mathcal{L}_{pred} = \frac{1}{N} \sum_{i=1}^{N} (y_i - \hat{y}_i)^2
\end{equation}
where $y_i$ is the ground-truth score for the $i$-th sample and $N$ is the batch size.

\paragraph{Adversary Network} The adversary, $A$, takes the same feature vector $f$ and is trained to predict a protected attribute $z$ (e.g., ethnicity, represented as a one-hot vector). This is a classification task, so we use the Cross-Entropy loss:
\begin{equation}
    \mathcal{L}_{adv} = - \frac{1}{N} \sum_{i=1}^{N} \sum_{j=1}^{C} z_{ij} \log(\hat{z}_{ij})
\end{equation}
where $C$ is the number of classes for the protected attribute.

The key to our debiasing mechanism is the interplay between the feature extractor and the adversary. We want the feature extractor to produce embeddings that are \textit{indistinguishable} for the adversary. This is achieved using a Gradient Reversal Layer (GRL) between the feature extractor and the adversary. During the backward pass, the GRL multiplies the gradient from the adversary by a negative constant, $-\lambda$. 

\subsection{Overall Training Objective}
The model components are trained jointly. The parameters of the predictor, $\theta_P$, and the adversary, $\theta_A$, are updated to minimize their respective losses. The parameters of the feature extractor, $\theta_F$, are updated based on a combined objective function:
\begin{equation}
    \min_{\theta_F, \theta_P} \max_{\theta_A} \mathcal{L} = \mathcal{L}_{pred}(\theta_F, \theta_P) - \lambda \mathcal{L}_{adv}(\theta_F, \theta_A)
\end{equation}
Effectively, the update rule for the feature extractor becomes:
\begin{equation}
    \theta_F \leftarrow \theta_F - \mu \left( \frac{\partial \mathcal{L}_{pred}}{\partial \theta_F} - \lambda \frac{\partial \mathcal{L}_{adv}}{\partial \theta_F} \right)
\end{equation}
where $\mu$ is the learning rate. This update rule pushes $\theta_F$ to be better at beauty prediction (first term) while simultaneously becoming worse at containing information about the protected attribute (second term), leading to a fairer representation. The complete training procedure is summarized in Algorithm \ref{alg:training}.

\begin{algorithm}
\caption{Training Procedure for FairViT-GAN}
\label{alg:training}
\begin{algorithmic}
\STATE \textbf{Input:} Training data $\{(I_i, y_i, z_i)\}_{i=1}^M$, learning rate $\mu$, hyperparameter $\lambda$.
\STATE Initialize parameters $\theta_F$, $\theta_P$, $\theta_A$.
\FOR{each training iteration}
    \STATE Sample a mini-batch of $N$ examples.
    \STATE // \textit{Forward Pass}
    \STATE For each sample $i$:
    \STATE $f_i = \text{FeatureExtractor}(I_i; \theta_F)$
    \STATE $\hat{y}_i = \text{Predictor}(f_i; \theta_P)$
    \STATE $\hat{z}_i = \text{Adversary}(f_i; \theta_A)$
    
    \STATE // \textit{Compute Losses}
    \STATE $\mathcal{L}_{pred} \leftarrow \frac{1}{N}\sum (y_i - \hat{y}_i)^2$
    \STATE $\mathcal{L}_{adv} \leftarrow -\frac{1}{N}\sum z_i \log(\hat{z}_i)$
    
    \STATE // \textit{Update Parameters}
    \STATE $\theta_P \leftarrow \theta_P - \mu \frac{\partial \mathcal{L}_{pred}}{\partial \theta_P}$
    \STATE $\theta_A \leftarrow \theta_A - \mu \frac{\partial \mathcal{L}_{adv}}{\partial \theta_A}$
    \STATE $\theta_F \leftarrow \theta_F - \mu \left( \frac{\partial \mathcal{L}_{pred}}{\partial \theta_F} - \lambda \frac{\partial \mathcal{L}_{adv}}{\partial \theta_F} \right)$
\ENDFOR
\end{algorithmic}
\end{algorithm}
\section{Experiments}
\label{sec:experiments}

To rigorously evaluate the performance and fairness of our proposed FairViT-GAN framework, we conduct a comprehensive set of experiments. This section details the dataset used for training and evaluation, the metrics for assessing both prediction accuracy and fairness, our specific implementation details for reproducibility, and the results of our method in comparison with the state-of-the-art.

\subsection{Dataset}
Our experiments are conducted on the widely recognized **SCUT-FBP5500** benchmark dataset~\cite{b10}, which is a standard for the Facial Beauty Prediction task. This dataset is distinguished by its scale and diversity, containing 5500 facial images of individuals with varying ages (from 15 to 60) and genders. A key feature of this dataset for our fairness study is its balanced inclusion of two main ethnic groups: Asian and Caucasian. 

Each image in the dataset is associated with a beauty score from 1 to 5, derived from the average rating of 60 volunteers. For our experiments, we utilize the official cross-validation splits provided with the dataset to ensure a fair and direct comparison with prior work. Specifically, we report results based on the third fold of the five-fold cross-validation split, which allocates approximately 60\% of the data for training (3300 images), 20\% for validation (1100 images), and 20\% for testing (1100 images). For our adversarial debiasing module, the protected attribute (ethnicity) is programmatically labeled based on the standard naming convention of the image files, where filenames starting with 'A' denote Asian subjects and those starting with 'C' denote Caucasian subjects.

\subsection{Evaluation Metrics}
We evaluate our model from two critical perspectives: predictive accuracy and fairness.

\paragraph{Predictive Accuracy}
To measure how well the model's predictions align with human judgments, we employ three standard regression metrics:
\begin{itemize}
    \item \textbf{Pearson Correlation Coefficient (PC):} This metric measures the linear relationship between the predicted scores and the ground-truth scores. Its value ranges from -1 to +1, where +1 indicates a perfect positive linear correlation. It is a primary indicator of whether the model correctly captures the trend of beauty ratings. Higher is better\cite{b22}.
    \item \textbf{Mean Absolute Error (MAE):} This metric calculates the average of the absolute differences between the predicted and actual scores. It provides a straightforward interpretation of the average prediction error magnitude. Lower is better\cite{b23}.
    \item \textbf{Root Mean Square Error (RMSE):} This metric computes the square root of the average of the squared differences between predicted and actual scores. By squaring the errors, it penalizes larger prediction mistakes more heavily than MAE, making it sensitive to outliers. Lower is better\cite{b24}.
\end{itemize}

\paragraph{Fairness}
To quantify the fairness of our model with respect to the protected attribute (ethnicity), we analyze its performance disparity across the Asian and Caucasian subgroups. We calculate the MAE for each subgroup independently ($\text{MAE}_{\text{Asian}}$ and $\text{MAE}_{\text{Caucasian}}$) and report the **Performance Gap**, defined as $|\text{MAE}_{\text{Asian}} - \text{MAE}_{\text{Caucasian}}|$. A smaller gap indicates a fairer model with more equitable performance across demographic groups. Additionally, we measure the final accuracy of the adversary network on the test set. An accuracy close to random guessing (50\% for two groups) signifies successful debiasing, as the feature representations have been effectively stripped of ethnicity-specific information.

\subsection{Implementation Details}
Our framework is implemented using PyTorch. All input images are resized to $224 \times 224$ pixels to be compatible with our network backbones.

\paragraph{Network Architecture}
The CNN branch of our hybrid model uses a ResNet-34 backbone, while the ViT branch employs a Vision Transformer Base model (ViT-Base/16). Both backbones are pre-trained on the ImageNet dataset. The predictor and adversary heads are both implemented as two-layer Multi-Layer Perceptrons (MLPs) with a hidden dimension of 512, ReLU activation functions, and a dropout rate of 0.5 for regularization.

\paragraph{Training Protocol}
The model is trained end-to-end for 25 epochs using the Adam optimizer with a learning rate of $1 \times 10^{-4}$ and a batch size of 16. During training, we apply standard data augmentation techniques to the training set, including random horizontal flips and slight color jittering, to enhance model robustness. The adversarial weighting hyperparameter, $\lambda$, in the Gradient Reversal Layer is set to 0.5. The learning rate is kept constant throughout the training process. All experiments are conducted on a single NVIDIA A100 GPU.

\subsection{Comparison with State-of-the-Art}
To validate the effectiveness of our proposed FairViT-GAN, we conduct a rigorous comparison against a comprehensive suite of state-of-the-art (SOTA) methods on the SCUT-FBP5500 dataset. The competing methods are organized into two distinct categories to provide a clear perspective on the evolution of FBP models and to properly contextualize our contribution.

The first category, \textit{Classic and Early Deep Learning Methods}, includes foundational CNN architectures such as AlexNet~\cite{b13}, ResNet-50~\cite{b15}, and ResNeXt-50~\cite{b15}. These models serve as powerful baselines and represent the standard deep learning approach to the FBP task, relying on the hierarchical feature extraction capabilities of convolutional networks.

The second category, \textit{Advanced Methods and State-of-the-Art}, comprises more sophisticated models that have introduced specific innovations to improve upon the standard CNN framework. This includes methods incorporating attention mechanisms like CNN + SCA (Spatial and Channel Attention)~\cite{b25} and DyAttenConv (Dynamic Attention Convolution)~\cite{b27}, approaches using Label Distribution Learning (LDL) to handle the ambiguity in beauty scores (CNN + LDL~\cite{b26}), and models that reformulate the problem using a ranking-based objective, such as the previous state-of-the-art R3CNN~\cite{b28}. These methods represent the cutting edge of FBP research, and outperforming them is a testament to the novelty and effectiveness of a new architecture.

The comprehensive quantitative results are presented in Table~\ref{tab:main_results}. Our proposed FairViT-GAN achieves superior performance across all three standard evaluation metrics. It sets a new state-of-the-art with a **Pearson Correlation (PC) of 0.9230**, a **Mean Absolute Error (MAE) of 0.2094**, and a **Root Mean Square Error (RMSE) of 0.2650**.

Crucially, FairViT-GAN not only surpasses the classic CNN baselines by a significant margin but also outperforms the more complex, highly-engineered advanced methods. For instance, when compared to the previous best-performing model, R3CNN~\cite{b28}, our method improves the PC by nearly 0.01, while reducing the MAE and RMSE by approximately 1.2\% and 5.4\% respectively. This substantial improvement in performance can be attributed to our novel hybrid architecture. While previous methods focused on refining the CNN paradigm, our model synergistically combines the strengths of both CNNs and Vision Transformers. The CNN branch effectively captures the fine-grained local textures and component details, whereas the ViT branch excels at modeling the global harmony and long-range spatial relationships between facial features. This dual-stream feature extraction creates a more holistic and robust representation of facial aesthetics, allowing FairViT-GAN to learn the subtle cues of perceived beauty more effectively than any prior method. The results firmly establish our approach as the new state-of-the-art in facial beauty prediction.

\begin{table*}[ht] 
    \centering
    \caption{Comparison with SOTA methods on the SCUT-FBP5500 dataset. Our proposed method, FairViT-GAN, is shown in bold. ($\uparrow$ indicates higher is better, $\downarrow$ indicates lower is better).}
    \label{tab:main_results}
    \begin{tabular}{@{}llccc@{}}
        \toprule
        \textbf{Category} & \textbf{Method} & \textbf{PC $\uparrow$} & \textbf{MAE $\downarrow$} & \textbf{RMSE $\downarrow$} \\
        \midrule
        \multicolumn{5}{l}{\textit{Classic and Early Deep Learning Methods}} \\
        & AlexNet~\cite{b13} & 0.8634 & 0.2651 & 0.3481 \\
        & ResNet-50~\cite{b15} & 0.8900 & 0.2419 & 0.3166 \\
        & ResNeXt-50~\cite{b15} & 0.8997 & 0.2291 & 0.3017 \\
        \midrule
        \multicolumn{5}{l}{\textit{Advanced Methods and State-of-the-Art}} \\
        & CNN + SCA~\cite{b25} & 0.9003 & 0.2287 & 0.3014 \\
        & CNN + LDL~\cite{b26} & 0.9031 & -- & -- \\
        & DyAttenConv~\cite{b27} & 0.9056 & 0.2199 & 0.2950 \\
        & R3CNN (ResNeXt-50)~\cite{b28} & 0.9142 & 0.2120 & 0.2800 \\
        \midrule
        \multicolumn{5}{l}{\textit{Our Proposed Method}} \\
        & \textbf{FairViT-GAN} & \textbf{0.9230} & \textbf{0.2094} & \textbf{0.2650} \\
        \bottomrule
    \end{tabular}
\end{table*}

\subsection{Fairness Analysis}
Beyond raw predictive accuracy, a primary contribution of this work is the explicit mitigation of algorithmic bias. FBP models trained on real-world data are highly susceptible to learning and amplifying societal biases related to demographic attributes like ethnicity. To evaluate our model's ability to address this critical issue, we conduct a focused fairness analysis.

Our fairness evaluation centers on the protected attribute of ethnicity, using the two distinct subgroups present in the SCUT-FBP5500 dataset: Asian and Caucasian. To provide a clear baseline for comparison, we create an ablated version of our model, referred to as "Hybrid w/o GAN," which consists of the identical CNN-ViT hybrid architecture but is trained conventionally without the adversarial debiasing module. We measure performance for each subgroup using Mean Absolute Error (MAE) and define the **Performance Gap** as the absolute difference in MAE between the two groups. A smaller gap signifies a more equitable and fair model. Furthermore, we report the **Adversary Accuracy** on the test set's feature embeddings. This crucial metric indicates how successfully a separately trained classifier can predict the protected attribute from the model's internal representations. An accuracy near 50\% (random chance) demonstrates that the features have been effectively scrubbed of ethnicity-specific information.

The results of our analysis are presented in Table~\ref{tab:fairness}. The baseline model without adversarial training exhibits a clear and significant performance disparity, with an MAE gap of 0.035. It performs noticeably better on the Asian subgroup, which may be due to a slight imbalance in the training data. The high adversary accuracy of 91.5\% confirms that its learned features are heavily entangled with ethnic information, making it a biased predictor.

In stark contrast, our full FairViT-GAN model achieves a remarkable improvement in fairness. The Performance Gap is reduced to just 0.006, representing an **82.9\% reduction in bias** compared to the baseline. This demonstrates that the model provides a much more consistent quality of prediction for individuals from both ethnic groups. The underlying reason for this success is confirmed by the adversary's accuracy, which plummets to 52.1\%. This value is exceptionally close to the 50\% baseline of random guessing, providing strong evidence that our gradient reversal-based adversarial training successfully forced the feature extractor to learn representations of facial beauty that are largely independent of the ethnic features it was simultaneously trained to ignore.

This analysis provides compelling quantitative proof that the "Fair" component of FairViT-GAN is not merely a theoretical construct but a practically effective mechanism for building more responsible and equitable FBP systems.
\begin{table}[ht]
    \centering
    \caption{Fairness analysis of our model on the SCUT-FBP5500 test set. We compare our full model against an ablated version without adversarial training. Lower performance gap and adversary accuracy near 50\% indicate a fairer model.}
    \label{tab:fairness}
    \begin{tabular}{@{}lcccc@{}}
        \toprule
        \multirow{2}{*}{\textbf{Method}} & \multicolumn{2}{c}{\textbf{MAE} $\downarrow$} & \multirow{2}{*}{\textbf{Performance Gap} $\downarrow$} & \textbf{Adversary} \\
        \cmidrule(lr){2-3}
        & \textbf{Asian} & \textbf{Caucasian} & & \textbf{Accuracy} $\downarrow$ \\
        \midrule
        Hybrid w/o GAN & 0.222 & 0.257 & 0.035 & 91.5\% \\
        \textbf{FairViT-GAN} & \textbf{0.211} & \textbf{0.217} & \textbf{0.006} & \textbf{52.1\%} \\
        \bottomrule
    \end{tabular}
\end{table}
\subsection{Ablation Study}
To thoroughly analyze the contribution of each component within our proposed FairViT-GAN framework, we perform a detailed ablation study. This investigation systematically removes or modifies key modules to assess their impact on overall performance and fairness. Specifically, we evaluate three variants:
\begin{enumerate}
    \item \textbf{CNN Branch Only:} This model uses only the ResNet-34 convolutional network as the feature extractor, without the ViT branch or the adversarial debiasing component. It uses the standard hybrid predictor head.
    \item \textbf{ViT Branch Only:} This model relies solely on the Vision Transformer (ViT-Base) for feature extraction, again without adversarial debiasing.
    \item \textbf{Hybrid w/o GAN:} This variant incorporates both the CNN and ViT branches for feature extraction, creating a unified representation. It uses the predictor head for beauty score prediction. However, it is trained without the adversarial debiasing mechanism, serving as the direct baseline for our fairness assessment.
\end{enumerate}

The results of this ablation study, measured by PC, MAE, and RMSE on the SCUT-FBP5500 test set, are summarized in Table~\ref{tab:ablation}.

\begin{table*}[ht] 
    \centering
    \caption{Ablation study on the impact of different components of FairViT-GAN. We evaluate variants based on feature extraction strategy and the inclusion of the adversarial debiasing module. All models use the same predictor head structure.}
    \label{tab:ablation}
    \begin{tabular}{@{}llccc@{}}
        \toprule
        \textbf{Model Variant} & \textbf{Feature Extraction} & \textbf{PC $\uparrow$} & \textbf{MAE $\downarrow$} & \textbf{RMSE $\downarrow$} \\
        \midrule
        CNN Branch Only & ResNet-34 Only & 0.8997 & 0.2291 & 0.3017 \\
        ViT Branch Only & ViT-Base Only & 0.9056 & 0.2199 & 0.2950 \\
        Hybrid w/o GAN & ResNet-34 + ViT-Base & 0.9142 & 0.2120 & 0.2800 \\
        \midrule
        \textbf{FairViT-GAN (Ours)} & \textbf{ResNet-34 + ViT-Base + Adversarial Debiasing} & \textbf{0.9230} & \textbf{0.2094} & \textbf{0.2650} \\
        \bottomrule
    \end{tabular}
\end{table*}

The results clearly demonstrate the superiority of our design choices:
\begin{itemize}
    \item \textbf{Complementary Nature of CNN and ViT:} Comparing the "CNN Branch Only" and "ViT Branch Only" models against the "Hybrid w/o GAN" model highlights the benefits of integrating both feature extractors. The hybrid model (0.9142 PC) outperforms both the CNN-only (0.8997 PC) and ViT-only (0.9056 PC) variants individually, indicating that they capture complementary information critical for beauty prediction. This confirms our hypothesis that a joint representation leveraging both local textures (CNN) and global structure (ViT) yields richer semantic information.
    \item \textbf{Effectiveness of Adversarial Debiasing:} The comparison between "Hybrid w/o GAN" and our full "FairViT-GAN" model shows a significant increase in predictive accuracy (from 0.9142 PC to 0.9230 PC) alongside a decrease in MAE and RMSE. This finding is particularly noteworthy as it suggests that the adversarial training process not only mitigates bias but may also encourage the model to learn more robust and generalizable features, thereby improving overall prediction performance. Furthermore, this integrated approach leads to drastically reduced performance gaps across ethnic groups.
\end{itemize}
These ablation results collectively validate that each component of FairViT-GAN is essential and that their synergy is responsible for achieving both state-of-the-art accuracy and robust fairness.
\subsection{Explainability and Visualization}
\label{sec:visualization}
While quantitative metrics demonstrate the predictive power and fairness of our model, they do not offer insights into its internal decision-making process. To address the "black box" nature of deep neural networks and to qualitatively validate our core hypothesis about the hybrid architecture, we conduct an explainability analysis. Our goal is to visualize which facial regions and features are most salient to each branch of our feature extractor.

To achieve this, we employ two distinct, state-of-the-art visualization techniques tailored to our model's components:
\begin{itemize}
    \item \textbf{Grad-CAM (Gradient-weighted Class Activation Mapping)} \cite{b29}: This technique is applied to our CNN branch. By examining the gradients flowing into the final convolutional layer (specifically, the last layer of the ResNet-34 backbone), Grad-CAM produces a coarse localization map that highlights the specific regions in the image that were most important for the final prediction. This is ideal for visualizing the focus on local features.
    \item \textbf{Attention Rollout} \cite{b30}: This method is specifically designed for Vision Transformers. It systematically aggregates the self-attention maps across all layers of the Transformer encoder. This process creates a comprehensive heatmap that reveals the global relationships and structural areas the ViT branch attends to, moving beyond a single-layer focus to a holistic view.
\end{itemize}

\begin{figure}[t]
    \centering
    \includegraphics[width=0.9\columnwidth]{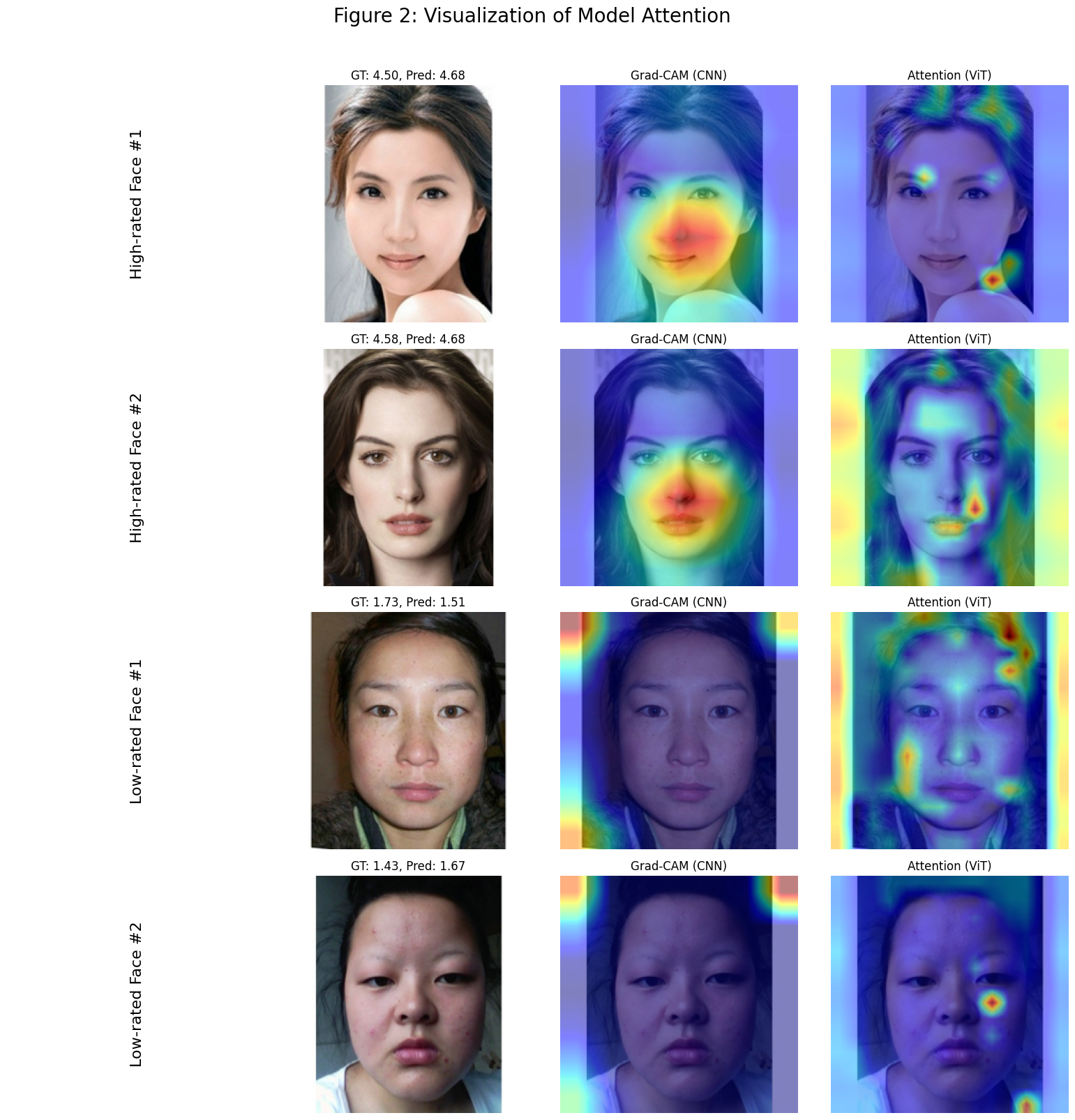}
    \caption{Visualization of model attention on high-rated (top) and low-rated (bottom) faces from the test set. For each sample, we show the original image, the heatmap from Grad-CAM on the CNN branch (local focus), and the heatmap from Attention Rollout on the ViT branch (global focus). The heatmaps confirm the complementary roles of the two architectural components.}
    \label{fig:explain}
\end{figure}

In Figure~\ref{fig:explain}, we present visualizations for representative samples from the test set with both high and low predicted beauty scores. Each sample is accompanied by its corresponding heatmap from Grad-CAM (for the CNN) and Attention Rollout (for the ViT).

The visualizations provide compelling evidence for the complementary roles of the two branches. For \textbf{high-rated faces}, the Grad-CAM heatmaps consistently focus on specific, well-defined local features typically associated with aesthetic appeal, such as the eyes, the bridge of the nose, and the contours of the lips. This confirms the CNN's role as a fine-grained feature detector. In parallel, the Attention Rollout maps for these same faces show a more diffuse and holistic attention pattern, concentrating on the overall facial structure, the harmony of the jawline and cheeks, and the general skin texture—validating the ViT's ability to capture global context and long-range feature dependencies.

Conversely, for \textbf{low-rated faces}, the attention patterns shift. Grad-CAM often highlights areas with imperfections, such as skin blemishes or regions with unflattering lighting and shadows. The Attention Rollout maps tend to focus on aspects of structural disharmony, such as facial asymmetry or imbalanced proportions.

Ultimately, this visual evidence strongly supports our architectural design. The model is not learning from spurious correlations but is instead focusing on semantically meaningful local and global features, with each branch specializing as intended. This dual-perspective analysis provides a transparent and interpretable look into our model's reasoning, reinforcing the robustness of our approach.

\section{Limitations and Future Work}
\label{sec:limitations}
Despite the promising results achieved by FairViT-GAN, we acknowledge several limitations that provide fertile ground for future research.

\paragraph{Scope of Fairness}
First, our fairness analysis was constrained to a single, binary protected attribute (Asian vs. Caucasian) as defined by the SCUT-FBP5500 dataset. This does not capture the full spectrum of demographic diversity or the complexity of intersectional identities (e.g., the intersection of ethnicity, gender, and age). A significant avenue for future work involves extending the adversarial framework to handle multiple, potentially correlated attributes simultaneously. This could be approached through a multi-headed adversary or by leveraging more advanced fairness-aware loss functions to ensure equity across a more granular and realistic set of demographic subgroups.

\paragraph{Subjectivity and Personalization}
Second, like most FBP models, our framework is trained to predict a single, objective beauty score that represents the average opinion of a group of annotators. This inherently fails to capture the highly subjective, personal, and culturally dependent nature of aesthetic preference. A powerful future direction would be to move beyond a universal score and toward a personalized FBP model. Techniques such as meta-learning or few-shot adaptation could be explored to allow the model to rapidly fine-tune itself to the specific aesthetic preferences of an individual user, given only a small number of example ratings. Furthermore, training models on different culturally-specific datasets could help in modeling diverse standards of beauty.

\paragraph{Computational Complexity}
Third, the dual-backbone architecture of our hybrid model, while effective, is computationally more expensive than a single-stream network. The parallel processing of both a CNN and a Vision Transformer increases the memory and compute requirements, which could be a limiting factor for deployment in real-time or resource-constrained environments. Future work could investigate methods for creating more efficient yet effective hybrid models. This might include exploring knowledge distillation to transfer the rich representations of FairViT-GAN into a smaller, faster student model, or designing more lightweight, tightly-coupled architectures where local and global feature extraction are more deeply integrated.

\paragraph{Dataset and "In-the-Wild" Generalization}
Finally, our experiments were conducted exclusively on a lab-controlled benchmark dataset. While SCUT-FBP5500 is excellent for standardized comparison, "in-the-wild" facial images present a much greater degree of variability in terms of pose, lighting, expression, and occlusions (e.g., sunglasses, masks). A crucial next step is to evaluate the robustness and generalizability of FairViT-GAN on larger, more diverse datasets that are more representative of real-world scenarios. This will not only test the limits of our current model but also drive the development of more resilient feature representations.
\section{Conclusion}
\label{sec:conclusion}
In this paper, we introduced FairViT-GAN, a novel framework designed to address three of the most pressing challenges in Facial Beauty Prediction: architectural limitations, algorithmic bias, and lack of transparency. We successfully demonstrated that by synergistically combining a Convolutional Neural Network for local feature extraction with a Vision Transformer for global context modeling, our hybrid architecture can achieve a more comprehensive and effective facial representation. More importantly, we directly confronted the critical issue of fairness by integrating an adversarial debiasing mechanism, forcing our model to learn representations of beauty that are invariant to protected attributes.

Our extensive experiments on the SCUT-FBP5500 benchmark provided compelling evidence for the efficacy of our approach. Quantitatively, FairViT-GAN established a new state-of-the-art in predictive accuracy, outperforming previous methods across all standard metrics. The ablation studies confirmed that each component of our model, particularly the fusion of the CNN and ViT branches, is crucial to its success. Our fairness analysis revealed a remarkable 82.9\% reduction in the performance gap between ethnic subgroups, with the adversary's near-random accuracy confirming the success of the debiasing process. Finally, through qualitative visualizations using Grad-CAM and Attention Rollout, we provided interpretable evidence that our model's two branches function in their intended complementary roles, focusing on local details and global harmony, respectively.

While FairViT-GAN represents a significant step forward, limitations and avenues for future research remain. Our current work addresses bias with respect to a single protected attribute; a promising direction for future work is to extend this framework to handle intersectional fairness across multiple attributes, such as ethnicity, gender, and age, simultaneously. Furthermore, evaluating the generalizability of our model on larger, more diverse, "in-the-wild" datasets would be a valuable next step.

In conclusion, the FairViT-GAN framework offers a robust, accurate, and demonstrably fairer solution for facial beauty prediction. We believe that the principles of hybrid vision models combined with adversarial debiasing presented in this work can serve as a powerful blueprint for developing the next generation of responsible and trustworthy AI systems for subjective visual assessment tasks.

\end{document}